%% file: acl_latex.tex
\pdfoutput=1

\documentclass[11pt]{article}

\usepackage[preprint]{acl}
\usepackage{float}
\usepackage[utf8]{inputenc}
\usepackage{times}
\usepackage{latexsym}
\usepackage{natbib}
\usepackage{booktabs}
\usepackage{multirow}
\usepackage{amsmath}
\usepackage[inkscapelatex=false]{svg}
\usepackage[T1]{fontenc}

\usepackage[utf8]{inputenc}

\usepackage{microtype}

\usepackage{inconsolata}
\usepackage{tabularx}
\usepackage{longtable}
\usepackage{cleveref}
\usepackage{array}
\usepackage{adjustbox}
\usepackage{color, soul}
\usepackage{pifont}
\usepackage{subcaption} 
\usepackage{CJKutf8}
\usepackage{graphicx}
\crefname{section}{§}{§§}
\Crefname{section}{§}{§§}
\usepackage[ruled]{algorithm2e}

\newcommand\ourmethod{\textsc{StoryWriter}\xspace}
\newcommand\ourmodelglm{\textsc{StoryWriter\textsubscript{GLM}}\xspace}
\newcommand\ourmodelllama{\textsc{StoryWriter\textsubscript{Llama}}\xspace}

\newcommand\ourdataset{\textsc{LongStory}\xspace}

\title{\ourmethod: A Multi-Agent Framework for Long Story Generation}

\author{Haotian Xia\thanks{\quad Equal contribution.}, Hao Peng$^{*}$, Yunjia Qi, Xiaozhi Wang, Bin Xu, Lei Hou, Juanzi Li\\Department of Computer Science and Technology, BNRist, Tsinghua University \\\texttt{\{xiaht24,peng-h24\}@mails.tsinghua.edu.cn}}

\begin{document}
\begin{CJK}{UTF8}{gbsn}
\maketitle
\begin{abstract}

Long story generation remains a challenge for existing large language models (LLMs), primarily due to two main factors: (1) discourse coherence, which requires plot consistency, logical coherence, and completeness in the long-form generation, and (2) narrative complexity, which requires an interwoven and engaging narrative.
To address these challenges, we propose \ourmethod, a multi-agent story generation framework, which consists of three main modules: (1) outline agent, which generates event-based outlines containing rich event plots, character, and event-event relationships. (2) planning agent, which further details events and plans which events should be written in each chapter to maintain an interwoven and engaging story. (3) writing agent, which dynamically compresses the story history based on the current event to generate and reflect new plots, ensuring the coherence of the generated story. We conduct both human and automated evaluation, and \ourmethod significantly outperforms existing story generation baselines in both story quality and length. 
Furthermore, we use \ourmethod to generate a dataset, which contains about $6,000$ high-quality long stories, with an average length of $8,000$ words. We train the model Llama3.1-8B and GLM4-9B using supervised fine-tuning on \ourdataset and develop \ourmodelllama and \ourmodelglm, which demonstrates advanced performance in long story generation. All code, models, and data are made publicly available\footnote{\url{https://github.com/THU-KEG/StoryWriter}} to encourage reproducibility and further development.





\end{abstract}

\input{01.intro}
\input{04.methods}
\input{05.evaluation}


\input{07.training}
\input{10.conclusion}

\input{08.limitations}
\input{11.ethical}

\bibliography{custom}

\end{CJK}
\end{document}

%% file: 01.intro.tex
\section{Introduction}
Story generation aims to automatically produce coherent, organized, and engaging narratives~\citep{wang2023open}. Typically, story generation involves using a premise, often a brief beginning or theme, as input to create a complete narrative~\citep{alhussain2021automatic}. Since the emergence
of large language models (LLMs; \citealp{ouyang2022training}), 
the quality of generated stories using LLMs has steadily improved~\citep{xie2024creating}. However, generating long stories, particularly those exceeding $1,000$ words, remains a significant challenge for LLMs~\citep{migal2024overview}.

The main challenges of long story generation are from two aspects: (1) discourse coherence, which requires plot consistency, logical coherence, and completeness in long-form generation. Existing LLMs still face challenges in generating fluent long texts~\citep{liu2024longgenbench}. In long story generation, LLMs need to retain long-distance key information, such as events, characters, and their relationships, to ensure plot consistency across the narrative. (2) narrative complexity, which requires interwoven, engaging, and diverse story content. While human-written stories typically exhibit these characteristics, LLM-generated narratives are often homogeneous, lacking in diversity and plot development~\citep{tian2024large,wang2024generating}.

\begin{figure}
    \centering
    \includegraphics[width=0.98\linewidth]{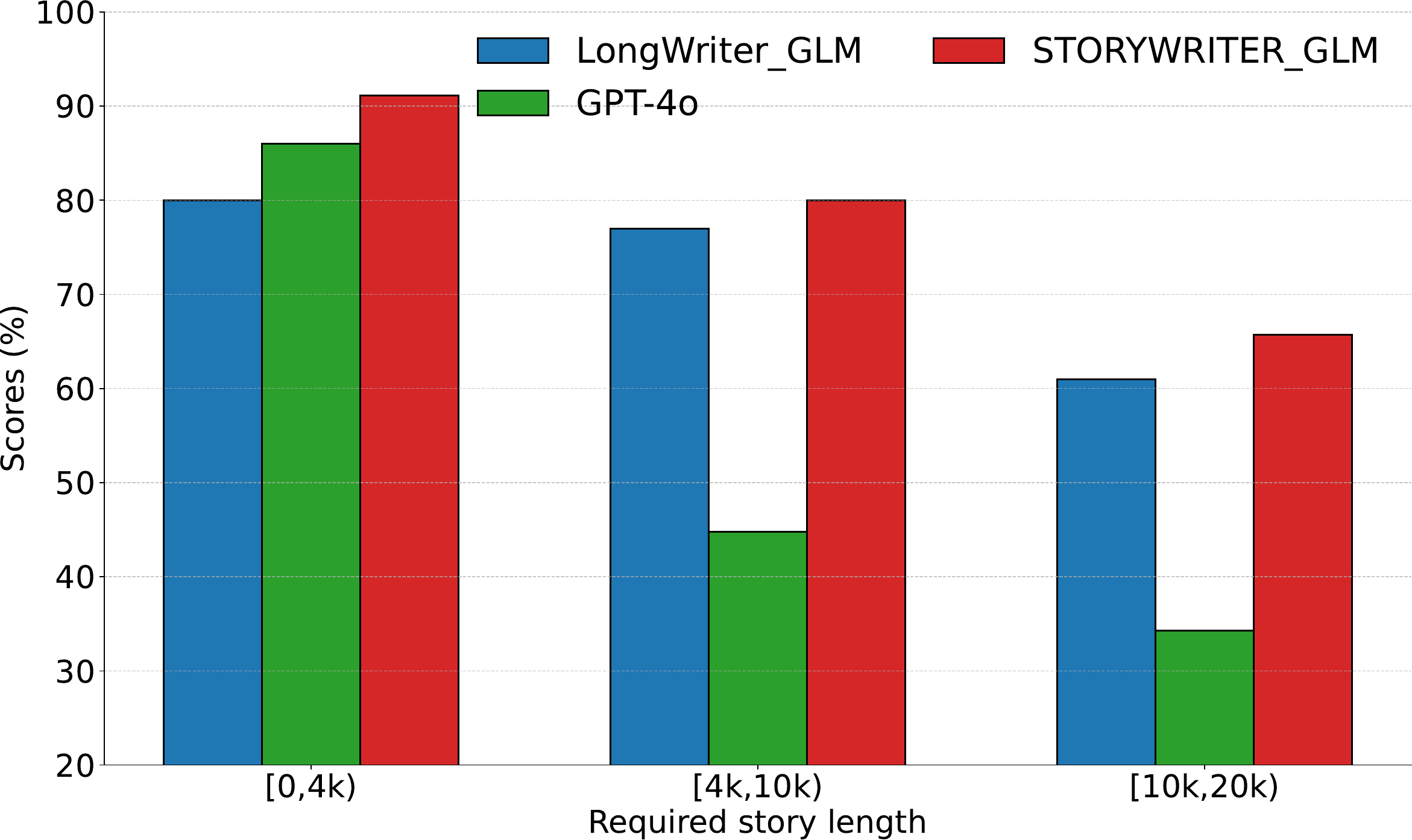}
    \caption{Results on MoPS~\citep{ma2024mopsmodularstorypremise} with different required story lengths. Details are placed in \cref{sec:dataset}.}
    \label{fig:enter-label}
\end{figure}


To address the above challenges, we propose \ourmethod, a multi-agent framework for long story generation, which consists of three main modules:
(1) \textbf{outline agent}, which generates event-based outlines. Generating outlines is a typical procedure in story generation, previous studies adopt LLMs to directly generate outlines~\citep{wang2023improving,yang2023doc,wang2024generating}, which may be insufficiently specific and diverse. Inspired by conventional event knowledge~\citep{wang2023maven}, we adopt an agent to generate a detailed event graph, where each node 
represents an event, and edges represent relationships between events, such as causal relationships~\citep{wang2022maven}. 
Each event is associated with several characters~\citep{wang2023maven}. We then adopt an agent to validate the consistency of each event and produce the final outline.
(2) \textbf{planning agent}, which generates detailed sub-events and globally plans which events should appear in each chapter to maintain an interwoven and engaging story. Specifically, we first use LLMs to generate sub-events for each event to provide richer event information. Human writing is non-linear, with events and characters often linked in diverse ways across different chapters~\citep{oller1983story,alkaaf2017tell}. We also employ an LLM to globally plan which events and characters should appear in each chapter, ensuring consistency and enabling the reappearance of key elements across chapters. This helps mitigate homogeneity and promotes the creation of interwoven content.
(3) \textbf{writing agent}, which generates and refines specific story content based on the historical context. Long story generation involves long-range dependencies and directly feeding the entire history to the LLM may result in missing key information~\citep{liu2024lost}, we adopt an agent named Coordinator to dynamically compress the previous writing history based on the current event. The goal of compression is to retain only relevant events and characters and create a compact and effective writing history for generating a more coherent story. We then input this history with an event requiring expansion to the final writer to generate a sub-story, and then refine it using the Coordinator.

We conduct extensive experiments to validate the effectiveness of \ourmethod. We adopt GPT-4o-mini~\citep{OpenAI2024} as the backbone to implement \ourmethod. We conduct evaluation on the widely used MoPS dataset~\citep{ma2024mopsmodularstorypremise}. We also investigate several strong baselines, including DOC~\citep{yang2023docimprovinglongstory}, Agents' Room~\citep{huot2024agentsroomnarrativegeneration}, and GPT-4o-mini~\citep{OpenAI2024}. We adopt both human evaluation and GPT-4o-based automated evaluation across $6$ commonly used dimensions\citep{chhun2024languagemodelsenjoystories}, including relevance, coherence, empathy, surprise, creativity, and complexity. \ourmethod significantly outperforms other models, demonstrating its effectiveness. Additionally, we perform ablation studies on different modules and find that removing any module leads to a considerable decline in performance, which further demonstrates the importance and efficacy of each module.
Finally, we adopt \ourmethod to generate a training dataset, \ourdataset, which contains about $6,000$ stories with an average length of $15,000$ words. We fine-tune the Llama3.1-8B Instruct model~\citep{dubey2024llama} using supervised fine-tuning on \ourdataset to develop \ourmodelllama. We evaluate the trained model using LongWriter-Ruler and LongBench-Write~\citep{bai2024longwriterunleashing10000word}, and find that \ourmodelllama significantly outperforms Llama3.1-8B Instruct on story exceeding $2,000$ words, and even surpasses GPT-4o~\citep{OpenAI20244o}. This demonstrates the effectiveness of \ourdataset.

In conclusion, our contributions are mainly threefold:
(1) We propose \ourmethod, a multi-agent framework for generating high-quality long story.
(2) We construct a high-quality long story dataset \ourdataset. The dataset can be used for evaluation and training in the field of story generation.
(3) We conduct extensive experiments to demonstrate the effectiveness of \ourmethod, from which we develop an advanced LLM \ourmodelllama and \ourmodelglm for long story generation.

%% file: 04.methods.tex
\begin{figure*}[t]
  \includegraphics[width=\linewidth]{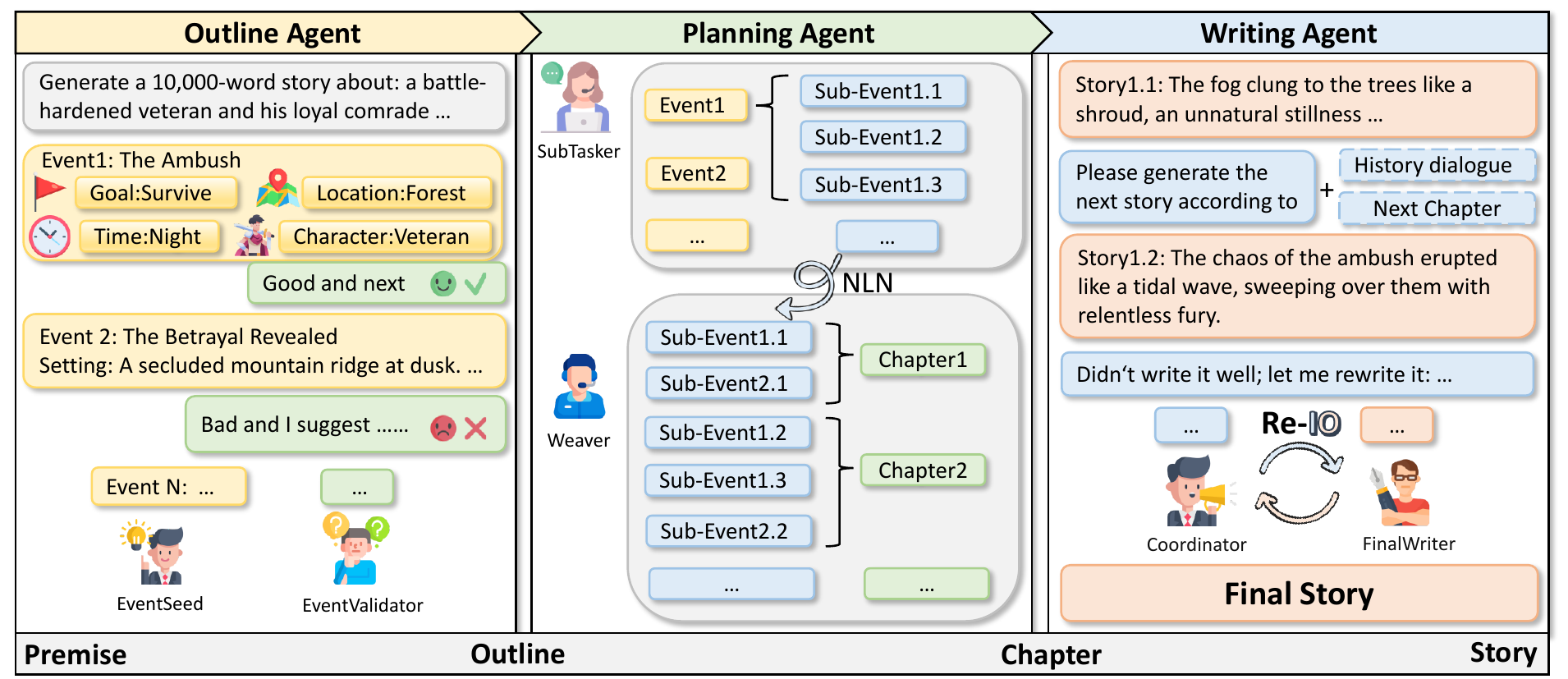}
  \caption {An overview of the three-stage story generation framework. The process consists of (from left to right): (1) event-based outline generation by the Outline Agent, (2) chapter construction using Non-Linear-Narration (NLN) by the Planning Agent, and (3) final story synthesis via ReIO (Re-write Input and Output) by the Writing Agent. Each stage employs a distinct methodology to progressively refine the narrative from high-level structure to detailed, coherent story text.}
  \label{fig:main_method}
\end{figure*}

\section{\ourmethod}
\subsection{Agents Net}
All components of \ourmethod are implemented within the Auto-Gen framework~\citep{wu2023autogenenablingnextgenllm}. The agent network is composed of three principal modules: outline agents, planning agents, and writing agents. The outline agents are responsible for generating the initial event-based outline, the planning agents refine and expand the outline into detailed sub-events and narrative structures, and the writing agents synthesize the final narrative text. By orchestrating multiple specialized agents for distinct roles, we establish a collaborative multi-agent writing paradigm (shown in Figure~\ref{fig:main_method}).
\subsection{Outline Agents}
For event-centric outline generation, our framework employs two specialized agents: \textit{EventSeed} and \textit{EventValidator}. The \textit{EventSeed} agent generates events sequentially based on the given premise, incrementally constructing the story outline by providing essential information for each event, such as time, location, and relationships. Meanwhile, the \textit{EventValidator} agent continuously monitors and evaluates the generated outline, offering feedback to ensure that each event is both plausible and narratively coherent, and guiding the generation of subsequent events. Distinct from conventional outline generation approaches that produce descriptive sentences, our method structures the outline as a sequence of event tuples, thereby enhancing both controllability and logical consistency.
\subsection{Planning Agents}

Enhancing flexibility and engagement in automated narrative generation remains a significant challenge. To address this, we introduce a novel Non-Linear Narration (NLN) strategy that decomposes events into sub-events and strategically distributes them across chapters. Grounded in Genette’s narrative order theory~\citep{genette1972narrative}, which distinguishes between story order and narrative order, our approach leverages techniques such as analepsis and prolepsis to enable complex, non-linear structures. Event structure and plot organization theories further underscore that narrative coherence depends on preserving causal and logical relationships among events~\cite{herman2002story, herman2017narrative}. As long as these links are maintained, readers can reconstruct the event chain, ensuring consistency even when sub-events are presented out of sequence. Additionally, Ryan’s ``narrative possible worlds'' framework~\citep{genette1980} highlights the potential for non-linear narratives to create diverse and interactive story paths. 

Building on these foundations, our NLN method systematically preserves the logical and causal integrity of events throughout decomposition and reorganization. Specifically, the SubTasker module is responsible for generating sub-events by decomposing high-level events into finer-grained narrative units. Subsequently, the Weaver module allocates these sub-events to different chapters, ensuring that the overall narrative structure remains coherent while enabling non-linear presentation. This division of labor allows for both detailed event modeling and flexible narrative organization, which are essential for implementing the NLN strategy. Even when sub-events are intentionally presented in a non-chronological order across chapters, the overall narrative coherence is preserved. This not only prevents narrative disruption or logical inconsistency but also endows the story with greater structural flexibility and expressive power, overcoming the monotony of linear narration and enhancing both narrative diversity and reader engagement.
\subsection{Writing Agents}
In the final generation phase, the collaborative interaction between the \textit{Coordinator} and \textit{FinalWriter} agents is pivotal for producing narratives that are both coherent in structure and consistent in style. The \textit{Coordinator} agent assumes responsibility for overseeing the global narrative architecture, engaging in all stages from outline formulation and sub-event planning to the ultimate text generation. In contrast, the \textit{FinalWriter} agent is primarily dedicated to synthesizing the final narrative, with a particular emphasis on ensuring stylistic uniformity and high textual quality. This division of labor ensures that both macro-level structural coherence and micro-level narrative fluency are achieved.
Despite these collaborative efforts, recent studies~\citep{yao2024sirllmstreaminginfiniteretentive} and our preliminary experiments have identified a critical challenge in long-form story generation: large language models (LLMs) exhibit significant context fragmentation and attention degradation when processing extended input sequences. Specifically, when the input length exceeds approximately 10,000 characters, the model’s capacity to maintain narrative focus and recall earlier plot developments diminishes substantially, frequently resulting in off-topic or incoherent outputs. This limitation presents a substantial obstacle to the generation of lengthy, cohesive narratives.

To address this issue, we propose the  Re-write Input and Output(ReIO) mechanism within the writing agents. During input processing, the \textit{Coordinator} dynamically summarizes and condenses the historical narrative context, selectively retaining only information pertinent to the current sub-event. This strategy effectively reduces the input length while preserving essential contextual information, and the generated summaries are cached for efficient reuse in subsequent stages. During output processing, the \textit{Coordinator} evaluates the generated text and, if necessary, rewrites it to ensure alignment with the intended narrative structure and stylistic requirements. The revised output replaces the original, and this iterative rewriting process is repeated as needed to maintain both narrative coherence and stylistic consistency.

By integrating the ReIO mechanism into the collaborative workflow of the \textit{Coordinator} and \textit{FinalWriter} agents, our framework effectively mitigates the challenges associated with long-context processing in LLMs, thereby enabling the generation of extended narratives that are both structurally robust and narratively engaging. For a detailed analysis of different history compression strategies, please refer to Section~\ref{sec:analysis_rewrite_context}.








%% file: 05.evaluation.tex
\input{table/main_experiment}
\input{table/ablation}

\section{Experiments}
\subsection{Experimental Setup}
\label{sec:exp_setup}
\paragraph{Evaluation Datasets} We use the dataset MoPS~\citep{ma2024mopsmodularstorypremise}.
They provide the MoPS code suite, along with 7.6k generated premises and 1,000 extended stories. Compared to premises generated by conventional methods and those collected from literary forums like WRITINGPROMPTS~\citep{fan-etal-2019-strategies}, the stories generated by MoPS exhibit higher quality and greater information density.

\paragraph{Evaluation Setup}
We adopt the evaluation framework proposed by HANNA~\citep{chhun-etal-2022-human}, a benchmark for story assessment, with minor adaptations to certain dimension definitions. This framework specifies six orthogonal criteria—Relevance, Coherence, Empathy, Surprise, Creativity, and Complexity—each grounded in social science literature. To comprehensively assess the generated stories, we employ both human and automated evaluation. For human evaluation, anonymized outputs are distributed to graduate students in an English program (all with TOEFL scores of 108 or above), who rate the stories on a five-point Likert scale across the six dimensions. For automated evaluation, we utilize GPT-4o~\citep{OpenAI20244o}, which assigns integer scores from 1 to 5 for each dimension. This dual evaluation protocol ensures a robust and multifaceted assessment of narrative quality.

\paragraph{Baselines} We compare stories generated by two methods DOC~\citep{yang2023docimprovinglongstory} and Agents' Room~\citep{huot2024agentsroomnarrativegeneration}:

(1) \textbf{DOC}. A method designed to enhance text quality by generating more comprehensive outlines.
For a fair comparison, we implemented the latest version of DOC’s methodology, using GPT-4o-mini as its base model. Instead of employing their automatic premise generation method, we directly utilized the premises provided in~\citet{ma2024mopsmodularstorypremise}. Additionally, due to factors such as API configuration changes over time, we made minor modifications to the underlying code of DOC while preserving its core logic.
(2) \textbf{Agents' Room}. Agents' Room is a multi-agent framework for story generation. This approach introduces an orchestrator responsible for determining when to invoke the writer agent and the planner agent, thereby ensuring coordinated execution among agents. However, experimental results in the original work indicate that, under the given experimental settings, the most effective strategy is a deterministic orchestrator that sequentially calls the agents in a predefined order. Accordingly, for consistency and comparability, we also adopted this deterministic orchestrator in our experiments.
(3) \textbf{GPT-4o mini}. We directly input the premise into GPT-4o-mini to generate the story.


\subsection{Experimental Results}
\paragraph{Main Results}
All the experimental results are presented in Table~\ref{table:main_experiment}. We observe the following:
(1) In general, our story generation framework \ourmethod significantly outperforms the baselines in both human and automated evaluations, demonstrating its effectiveness. (2) \ourmethod significantly surpasses previous baselines in terms of length while maintaining high generation quality, indicating its effectiveness in generating longer stories. (3) Across different specific evaluation dimensions, our method outperforms DOC and GPT-4o-mini in relevance and coherence, slightly falling behind Agents' Room. This may be due to that \ourmethod generates longer stories, and coherence inevitably decreases with increased length~\citep{bai2024longwriterunleashing10000word}. However, in terms of content diversity and creativity, our model significantly outperforms all baselines, validating the effectiveness of our approach and demonstrating that it can generate higher-quality, creative content, which is the ultimate goal of story generation.


\paragraph{Ablation Study}
The results of the ablation experiment are presented in Table~\ref{table:ablation}. We analyze the impact of removing key components from \ourmethod as follows:
\textbf{(-Events-Outline)}: This ablation removes event-based outlining, reducing the story outline to a few generic sentences without detailed event descriptions. In this case, the story outline lacks depth and structure, negatively impacting the quality of the generated stories. As a result, all six evaluation criteria show a significant decline, highlighting the importance of structured event-based outlines.
\textbf{(-Planning)}: This configuration eliminates the Non-Linear Narration (NLN) strategy in Planning Agents, causing sub-events to be arranged strictly in chronological order. As a result, the complexity score decreases significantly, second only to the (-Events-Outline) scenario. This is expected, as the Planning Agents module enhances narrative diversity by distributing sub-events across different chapters while preserving event relationships.
\textbf{(-ReIO-Input)}: In this setting, ReIO-input of Writing Agents is removed, meaning neither the input nor output is effectively regulated. Consequently, the input length for the agent increases substantially, leading to higher computational costs and a decline in overall performance.
\textbf{(-ReIO-Output)}: This setting removes the ReIO output rewriting mechanism in Writing Agents. In this case, the relevance score of the generated text drops significantly. This decline occurs because the ReIO output module plays a crucial role in maintaining structural coherence by rewriting sections that deviate from the original outline.


\subsection{Analysis on Summary Context}
\label{sec:analysis_rewrite_context}
As the length of generated text increases, LLMs are prone to undesirable phenomena such as repetition, hallucination, and topic drift~\citep{liu2024lost}. These issues typically manifest as redundant event narration, protagonist actions that diverge from the established narrative trajectory, and a breakdown in logical story progression relative to prior content. Our analysis reveals a strong correlation between these problems and the length of the preceding context. To mitigate these effects, we introduce a summary agent that condenses the input context while preserving essential information. Specifically, we implement a sliding window mechanism: as events are generated sequentially, the window advances, and the content within its range is systematically simplified.

A critical aspect of this approach is determining a strategy that optimally balances input length reduction with the preservation of narrative coherence. Through empirical evaluation of various window lengths, we observe that for texts under 15,000 tokens, a sliding window spanning [2, k-1] consistently yields optimal results, indicating that simplifying the central portion of the context is most effective.    

\begin{figure}[t]
  \includegraphics[width=\linewidth]{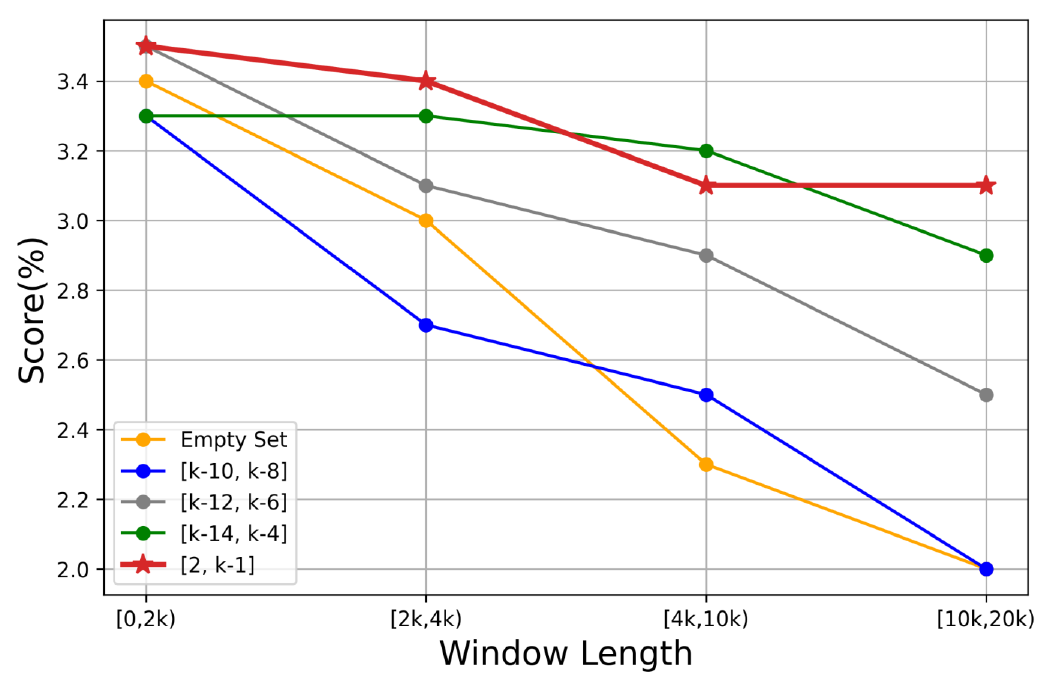}
  \caption {Results for different window lengths. The star ($^\star$) denotes the method with the best average performance across all cases.}
  \label{fig:window_length_table}
\end{figure}

To further substantiate our findings, we conduct a controlled experiment comparing five sliding window configurations: [k-10, k-8], [k-12, k-6], [k-14, k-4], the baseline [2, k-1], and an empty set. Human evaluators assess the narrative quality of the generated stories, with results presented in Figure~\ref{fig:window_length_table}. Our findings demonstrate that maximal simplification of prior content leads to superior narrative outcomes, as indicated by the best average performance across all cases (denoted by star in Figure~\ref{fig:window_length_table}).


%% file: table/main_experiment.tex
\begin{table*}[t]
\small
\centering

\begin{tabular}{llcccccccc}
\toprule
\multicolumn{1}{l}{Model}                          & \multicolumn{1}{l}{}     & \multicolumn{1}{c}{Average}                    & \multicolumn{1}{c}{RE} & \multicolumn{1}{c}{CH} &\multicolumn{1}{c}{EM} &\multicolumn{1}{c}{SU} &\multicolumn{1}{c}{CR} &\multicolumn{1}{c}{CX}   &\multicolumn{1}{c}{Average Length}                                   
\\ \midrule
\multirow{2}{*}{\textbf{DOC}} & Human-Eval & $3.7$ & $4.2$ & $4.3$ & $3.2$ & $3.4$ & $3.7$ & $3.2$  & \multirow{2}{*}{$2,373$} \\
                     & Auto-Eval & $3.9$ & $4.1$ & $4.3$ & $4.0$ & $3.5$ & $3.8$ & $3.5$ \\ 
                     \midrule
\multirow{2}{*}{\textbf{Agents' Room}} & Human-Eval & $3.8$ & $\boldsymbol{4.5}$ & $\boldsymbol{4.4}$ & $3.3$ & $3.2$ & $3.7$ & $4.0$ & \multirow{2}{*}{$3,134$} \\
                              & Auto-Eval & $3.9$ & $3.5$ & $4.5$ & $4.0$ & $3.7$ & $3.9$ &$3.7$          \\
                    \midrule
\multirow{2}{*}{\textbf{GPT-4o mini}} & Human-Eval & $3.6$ & $4.0$ & $3.8$ & $3.3$ & $3.4$ & $3.6$ & $3.7$ & \multirow{2}{*}{$1,078$}  \\
                           & Auto-Eval & $3.9$ & $4.0$ & $\underline{4.7}$ & $4.1$ & $3.5$ &$3.7$ & $3.4$ \\         
                           \midrule
\multirow{2}{*}{\textbf{\ourmethod}} & Human-Eval & $\boldsymbol{4.2}$ & $4.4$ & $4.3$ & $\boldsymbol{3.8}$ & $\boldsymbol{3.6}$ & $\boldsymbol{4.3}$ & $\boldsymbol{4.8}$ & \multirow{2}{*}{$8,081$}  \\
                           & Auto-Eval & $\underline{4.2}$ & $\underline{4.1}$ & $4.4$ & $\underline{4.4}$ & $\underline{3.7}$ & $\underline{4.2}$ & $\underline{4.6}$ \\    
\bottomrule
\end{tabular}
\caption{\label{table:main_experiment} Experimental results of human and automatic scoring (on a scale from 1 to 5). RE, CH, EM, SU, CR, and CX represent relevance, coherence, empathy, surprise, creativity, and complexity, respectively. \textbf{Bold} indicates the best result according to human evaluation, and \underline{underline} indicates the best result according to automatic evaluation.
}
\end{table*}


%% file: table/ablation.tex
\begin{table*}[t]
\centering
\small{
\begin{tabular}{lccccccc}
\toprule
\multicolumn{1}{l}{Model}       & \multicolumn{1}{c}{Average}                    & \multicolumn{1}{c}{RE} & \multicolumn{1}{c}{CH} &\multicolumn{1}{c}{EM} &\multicolumn{1}{c}{SU} &\multicolumn{1}{c}{CR} &\multicolumn{1}{c}{CX}                   
\\ \midrule
\multirow{1}{*}{\textbf{\ourmethod}} & $\textbf{4.3}$ & $\textbf{4.1}$ & $4.4$ & $4.4$ & $3.7$ & $\textbf{4.2}$ & $\textbf{4.6}$\\ 
\multirow{1}{*}{\quad (-) Events-Outlines}  & $2.5$ & $2.2$ & $3.2$ & $2.9$ & $2.2$ & $3.3$ & $1.1$       \\
\multirow{1}{*}{\quad (-) Planning} & $3.9$ & $4.0$ & $\textbf{4.6}$ & $4.0$ & $3.1$ & $3.9$ &$3.8$                      \\
\multirow{1}{*}{\quad (-) ReIO-Input} & $3.9$ & $4.1$ & $4.6$ & $3.9$ & $3.2$ & $3.9$ & $3.9$       \\ 
\multirow{1}{*}{\quad (-) ReIO-Output} & $4.0$ & $3.7$ & $4.2$ & $\textbf{4.6}$ & $\textbf{4.0}$ & $3.7$ & $3.9$\\
\bottomrule
\end{tabular}
}
\caption{\label{table:ablation}``(-)ReIO-Output'' removes the output rewriting mechanism of Writing Agents. (-)Planning removes the Non-Linear Narration (NLN) strategy of Planning Agents. (-)ReIO-Input removes ReIO input rewriting mechanismof Writing Agents. (-)Events-Outline removes event-based outlining of the Outline Agents, reducing the story outline to a few generic sentences without detailed event descriptions.~\textbf{Bold} indicates the best result according to auto evaluation}
\end{table*}

%% file: 07.training.tex
\section{Constructing \ourdataset}
\label{sec:dataset}
\input{table/sft_experiment}
\looseness=-1
In this section, we use \ourmethod to generate a high-quality long story dataset \ourdataset. We train the model Llama3.1-8B and GLM4-9B using supervised fine-tuning on \ourdataset and develop advanced storytelling LLM \ourmodelllama and \ourmodelglm. 
Our dataset shows significant improvement in doing sft on multiple downstream models.

\paragraph{\ourdataset Construction}

\looseness=-1
We construct a high-quality dataset with $5,500$ long-form stories, \ourdataset, using \ourmethod. Specifically, we first collect $6,000$ story premises from the training set of MoPS~\citep{ma2024mopsmodularstorypremise} and use \ourmethod to generate a long story for each premise. We then perform careful data cleaning to remove stories that are too short, do not meet format requirements, or exhibit low quality. Specifically, we merge multiple chapters of stories to mitigate the risk of overfitting to specific text structures during SFT training. As a result, we curate a final dataset comprising $5,500$ long stories, \ourdataset, with an average length of about $8, 000$ words.

\paragraph{Experimental Setup}
We adopt the same evaluation dataset MoPS in \cref{sec:exp_setup}. Due to the high cost of the manual evaluation, we only employ automated evaluation, which is also widely used in previous work~\citep{bai2024longwriterunleashing10000word,gu2024survey}. In addition to evaluating the content quality from $6$ dimensions mentioned in \cref{sec:exp_setup}, we also report the length score used by the LongBench-Write evaluation method~\citep{bai2024longwriterunleashing10000word}. This method controls the length of text generated by LLMs by setting different output length constraints, which not only assesses the model's ability to generate long texts but also evaluates its adherence to word count constraints. The length score computes the degree of alignment between the actual response length and the required length in the instruction, which can be computed as follows:

\begin{equation}
\label{eq:eq_sl}
    S_l = 
    \begin{cases} 
    100 \cdot \max \left( 0, 1 - \frac{(l' / l - 1)}{3} \right) & \text{if } l' > l, \\
    100 \cdot \max \left( 0, 1 - \frac{(l / l' - 1)}{2} \right) & \text{if } l' \leq l.
    \end{cases}
\end{equation}

$l'$ denotes the actual response length and $l$ denotes the required length. Specifically, we adopt the same evaluation settings as LongBench-Write: for each instruction in the MoPS test set, we add an output length constraint from \{$500$, $1,000$, $2,000$, $4,000$, $10,000$\}, and then generate response for each length constraint and compute the final scores. We bucket the results based on lengths and report the average of the following metrics within each bucket: $S_q$, which evaluates content quality (\textbf{the average of the $6$ dimensional scores} from \cref{sec:exp_setup}), $S_l$, which evaluates the length score, and $\bar{S}$, which equals $(S_q + 20 * S_l) / 2$. We also report the average overall score across all lengths.



\paragraph{SFT Training}
\looseness=-1
We leverage the Llama 3.1-8B model and GLM-4-9B model as the base model for SFT training. We use the training code proposed by LongAlign~\citep{bai2024longalign}, as it is specifically designed for long-context training with pre-existing long-context adaptations. We use the premise of each instance in \ourdataset as the input and the story as the output for supervised fine-tuning to obtain \ourmodelllama and \ourmodelglm, For both of which we set the batch size to $1$, learning rate to $2\times10^{-5}$, training $4$ epochs.




\paragraph{Experimental Results}
The experimental results of \ourmodelllama and \ourmodelglm trained on \ourdataset, along with other baselines, are shown in Table~\ref{table:performance_of_sft}. We can observe that: (1) For the quality of generated stories ($S_q$), \ourmodelglm significantly outperforms the backbone model, especially in generating stories over $4,000$ words. This indicates that \ourmodelglm can maintain high quality while generating longer content.
(2) For length scoring of the generated stories ($S_l$), our models also perform much better than Llama3.1-8B-Instruct and GPT-4o. This indicates that our models better adhere to length constraints in story generation. Although the training process does not involve explicit ability enhancement for following length constraints. This suggests that training with longer responses could enhance the model's ability to follow length constraints. In conclusion, \ourmodelllama and \ourmodelglm perform better in generating longer stories and adhering to length constraints, demonstrating the effectiveness of our data construction method \ourmethod and \ourdataset. As our approach can be extended to the broader field of creative content generation, we encourage the community to utilize it for producing more high-quality data.

%% file: table/sft_experiment.tex
\begin{table*}[t]
    \centering
    \small
    \renewcommand{\arraystretch}{1.2} %
    
    \resizebox{\linewidth}{!}{
    \begin{tabular}{lccccccccccccc}
        \toprule
        & \multicolumn{3}{c}{Overall} & \multicolumn{2}{c}{[0, 1k)} & \multicolumn{2}{c}{[1k, 2k)} & \multicolumn{2}{c}{[2k, 4k)} & \multicolumn{2}{c}{[4k, 10k)} & \multicolumn{2}{c}{[10k, 20k)} \\
        & $\bar{S}$ & $S_l$ & $S_q$ & $S_l$ & $S_q$ & $S_l$ & $S_q$ & $S_l$ & $S_q$ & $S_l$ & $S_q$ & $S_l$ & $S_q$ \\
        \midrule
        \textbf{Llama3.1-8B-Instruct} & $46.6$ & $34.5$ & $2.9$  & $89.0$ & $4.0$ & $83.7$ & $3.9$ & $0.0$ & $3.5$ & $0.0$ & $2.2$ & $0.0$ & $1.0$\\
        \textbf{GLM4-9B} & $47.3$ & $36.6$ & $2.9$  & $93.7$ & $4.2$ & $89.6$ & $4.0$ & $0.0$ & $3.3$ & $0.0$ & $2.0$ & $0.0$ & $1.0$\\
        \textbf{LongWriter-GLM4-9B} & $76.3$ & $83.0$ & $3.5$ & $86.9$ & $3.1$ & $93.1$ & $3.2$ & ${91.6}$ & $4.0$ & $86.9$ & $3.6$ & $56.7$ & $3.4$ \\
        \textbf{LongWriter-Llama3.1-8B} & $77.9$ & $83.6$ & $3.6$ & $96.9$ & $3.9$ & $96.1$ & $3.5$ & ${93.2}$ & $4.1$ & $81.9$ & $3.5$ & $51.3$ & $3.2$ \\
        \textbf{Deepseek-Llama-8B} & $70.0$ & $73.6$ & $3.3$ & $92.3$ & $3.1$ & $91.9$ & $3.2$ & $88.2$ & $3.6$ & $83.2$ & $3.4$ & $12.3$ & $3.3$ \\
        \textbf{Deepseek-Llama-70B} & $74.3$ & $79.0$ & $3.5$ & $93.2$ & $3.3$ & $94.5$ & $3.4$ & $87.2$ & $4.0$ & $81.0$ & $3.5$ & $44.1$ & $3.2$ \\
        \midrule
        \textbf{GPT-4o} & ${67.4}$ & $52.8$ & $\boldsymbol{4.1}$ & $92.3$ & $\boldsymbol{4.7}$ & $91.7$ & $\boldsymbol{4.5}$ & $62.0$ & $\boldsymbol{4.3}$ & $15.3$ & $\boldsymbol{3.7}$ & $2.7$ & $3.3$\\
        \midrule
        \textbf{\ourmodelllama} & $73.4$ & $75.3$ & $3.5$ & $90.8$ & $3.9$ & ${94.1}$ & $3.8$ & ${77.3}$ & $3.5$ & $77.0$ &$3.4$ & $27.7$ & $3.4$ \\
        \textbf{\ourmodelglm} & $\boldsymbol{83.7}$ & $\boldsymbol{88.5}$ & $3.9$ & $\boldsymbol{99.5}$ & $4.4$ & $\boldsymbol{99.3}$ & $4.1$ & $\boldsymbol{98.0}$ & $4.0$ & $\boldsymbol{88.7}$ &$3.5$ & $\boldsymbol{57.3}$ & $\boldsymbol{3.6}$ \\
        \bottomrule
    \end{tabular}
    }
    \caption{
    Experimental results (\%) of \ourmodelllama, \ourmodelglm and the baselines. $S_q$ represents the average score of the 6 dimensions, as described in \cref{sec:exp_setup}. $S_l$ is the length score, calculated using Equation~\ref{eq:eq_sl}. $\bar{S}$ is computed as $(S_q + 20 \times S_l) / 2$, following the approach used by \citet{bai2024longwriterunleashing10000word}.
    }
    \label{table:performance_of_sft}
\end{table*}

%% file: 10.conclusion.tex
\section{Conclusion}
This paper presents \ourmethod, a multi-agent approach that generates outlines and long-enough stories automatically. Using \ourmethod, we generate a large number of diverse and high-quality stories. Human and automatic evaluations demonstrate that \ourmethod outperforms multiple baselines. Similarly, we create a high-quality dataset \ourdataset using \ourmethod. We also perform supervised fine-tuning based on \ourdataset and provide \ourmodelllama based on Llama3.1-8B and \ourmodelglm on GLM4-9B. We believe that \ourmethod will be helpful for the long story generation task of LLM, and future Auto Story Generation(ASG) tasks can be explored based on these data and \ourmodelllama. We hope to explore LLM's generation of long serial novels further, which requires more powerful long-story generation and understanding capabilities from LLMs.

%% file: 08.limitations.tex
\section*{Limitations}
The limitations of this work are mainly threefold:(1)~There are some more powerful models than GPT-4o-mini to choose from, but considering the limited economic cost, we only used GPT-4o-mini as our generative model and used the generated data to distill an 8b lightweight model. This is obviously something that can be optimized.(2)~This study focuses exclusively on English-language data. In future research, we aim to extend our approach to support multiple languages, increasing its applicability across diverse linguistic contexts.(3)~ Our research primarily concentrates on novel-like story generation, with limited exploration of diverse artistic styles. Future work could investigate other narrative forms, such as scripts, poetry, and prose, to broaden the stylistic versatility of generated content.

%% file: 11.ethical.tex
\section*{Ethical Considerations}
We discuss the ethical considerations here:
(1) Intellectual property. 
We have strictly adhered to the licenses of all utilized artifacts, including datasets, models, and code repositories. We will open-source code, \ourdataset, \ourmodelglm and \ourmodelllama under the MIT license\footnote{\url{https://opensource.org/license/mit}}.
(2) Intended use and potential risk control.
We propose \ourmethod, a multi-agent story generation framework designed to produce coherent and complex stories. Additionally, we construct \ourdataset dataset based on MoPS to enhance the model's ability to generate long stories. We trust that the original publisher has appropriately anonymized and sanitized the dataset. Furthermore, \ourmethod generates creative stories with artistic embellishments, rather than real stories, and therefore does not introduce additional ethical concerns.
(3) AI assistance. 
We have used ChatGPT to refine some sentences.

